\documentclass[journal]{IEEEtran}
\ifCLASSINFOpdf
\else
\fi

\usepackage{graphicx}
\usepackage{amsmath}
\usepackage{amssymb}
\usepackage{colortbl}
\usepackage{diagbox}
\usepackage{subeqnarray}
\usepackage{cases}
\usepackage{setspace}
\usepackage{tabularx}
\usepackage{booktabs}
\usepackage{multicol}
\usepackage{dcolumn}
\usepackage{subfigure}
\usepackage{float}
\usepackage{epsfig}
\usepackage{array}
\usepackage{epstopdf}
\usepackage{color}
\usepackage{xcolor}
\usepackage[nocompress]{cite}
\usepackage{psfrag}
\usepackage{cite}
\usepackage{multirow}
\usepackage{color}

\usepackage{algorithm}
\usepackage{algorithmic}
\newcommand{\etal}{\textit{et al}.}
\newcommand{\ie}{\textit{i}.\textit{e}.}
\newcommand{\eg}{\textit{e}.\textit{g}.}
\usepackage[bookmarks,colorlinks,linkcolor=red, anchorcolor=blue, citecolor=green]{hyperref}
\usepackage{url}
\begin{document}
%
\title{Deep Identity-Aware Transfer \\of Facial Attributes}
%
%
%

\author{Mu~Li,
        Wangmeng~Zuo,~\IEEEmembership{Senior Member,~IEEE}
        David~Zhang,~\IEEEmembership{Fellow,~IEEE}
        ~and~Jane~You,~\IEEEmembership{Member,~IEEE}
\thanks{This project is partially supported by the HK RGC/GRF grant (under no. PolyU 5313/12E and PolyU 152212/14E) and the National Natural Science Foundation of China (NSFC) under Grant No. 61671182.}
\thanks{Mu Li and Jane You are with the Department
of Computing, Hong Kong Polytechnic University, Hong Kong, e-mail: (csmuli@comp.polyu.edu.hk,csyjia@comp.polyu.edu.hk).}
\thanks{Wangmeng Zuo is with the School
of Computer Science and Technology, Harbin Institute of Technology, Harbin, China, e-mail: (cswmzuo@gmail.com).}
\thanks{David Zhang is with the School of Science and Engineering, The Chinese University of Hong Kong (Shenzhen), Shenzhen, China, e-mail: (csdzhang@comp.polyu.edu.hk).}
\thanks{Manuscript received XXX; revised XXX}}

%
%

\markboth{}%
{Shell \MakeLowercase{\textit{et al.}}: Bare Demo of IEEEtran.cls for IEEE Journals}
%



\maketitle

\begin{abstract}
This paper presents a Deep convolutional network model for Identity-Aware Transfer (DIAT) of facial attributes.
Given the source input image and the reference attribute, DIAT aims to generate a facial image that owns the reference attribute as well as keeps the same or similar identity to the input image.
In general, our model consists of a mask network and an attribute transform network which work in synergy to generate photo-realistic facial image with the reference attribute.
Considering that the reference attribute may be only related to some parts of the image, the mask network is introduced to avoid the incorrect editing on attribute irrelevant region.
Then the estimated mask is adopted to combine the input and transformed image for producing the transfer result.
For joint training of transform network and mask network, we incorporate the adversarial attribute loss, identity aware adaptive perceptual loss and VGG-FACE based identity loss.
Furthermore, a denoising network is presented to serve for perceptual regularization to suppress the artifacts in transfer result, while an attribute ratio regularization is introduced to constrain the size of attribute relevant region.
Our DIAT can provide a unified solution for several representative facial attribute transfer tasks, \eg, expression transfer, accessory removal, age progression and gender transfer, and can be extended for other face enhancement tasks such as face hallucination.
The experimental results validate the effectiveness of the proposed method.
Even for the identity-related attribute (\eg, gender), our DIAT can obtain visually impressive results by changing the attribute while retaining most identity-aware features.
\end{abstract}


%
\IEEEpeerreviewmaketitle

\section{Introduction}
\label{sec:intro}
Face attributes, \eg, gender and expression, can not only provide a natural description of facial images~\cite{kumar2009attribute}, but also offer a unified viewpoint for understanding many facial animation and manipulation tasks. For example, the goal of facial avatar~\cite{ichim2015avatar} and reenactment~\cite{thies2016face} is to transfer the facial expression attributes of a source actor to a target actor. In most applications such as expression transfer, accessory removal and age progression, the animation only modifies the related attribute without changing the identity. But for some other tasks, the change of some attributes, \eg, gender and ethnicity, will inevitably alter the identity of the source image.

In recent years, a variety of methods have been developed for specific facial attribute transfer tasks, and have achieved impressive results. For expression transfer, approaches have been suggested to create 3D or image-based avatars from hand-held video~\cite{ichim2015avatar}, while face trackers and expression modeling have been investigated for offline and online facial reenactment~\cite{kemelmacher-shlizerman2010, thies2016face}. For age progression, explicit and implicit synthesis methods have been proposed for different image models~\cite{fu2010age, Kemelmacher-Shlizerman2014age}. Hair style generation and replacement have also been studied in literatures~\cite{hu2015hair, kemelmacher-shlizerman2016hair}.

Convolutional neural network (CNN)-based models have also been investigated for human face generation with attributes. Kulkarni \etal~\cite{kulkarni2015deep} propose deep convolution inverse graphic network (DG-IGN). This method requires a large number of faces of a single person for training, and can only generate faces with different pose and light. Gauthier~\cite{gauthier2014conditional} developes a conditional generative adversarial network (cGAN) to generate facial image from a noise distribution and conditional attributes. Yan \etal~\cite{yan2015attribute2image} suggest an attribute-conditioned deep variational auto-encoder which extracts the latent variables from a reference image and combines them with attributes to produce the generated image with a generative model. Oord \etal~\cite{Oord2016decoders} propose a conditional image generation model based on PixelCNN decoder for image generation conditioned on an arbitrary feature vector. However, the identity of the generated face is not emphasized in~\cite{gauthier2014conditional, yan2015attribute2image, Oord2016decoders}, making them not directly applicable to attribute transfer.

Motivated by the strong capability of CNN in modeling complex transformation~\cite{johnson2016perceptual} and capturing perceptual similarity~\cite{gatys2015style}, several approaches have also been suggested for facial attribute transfer.
Li \etal~\cite{li2016convolutional} suggest a CNN-based attribute transfer model from the optimization perspective, but both run time and transfer quality is far from satisfying.
Considering that it is impracticable to collect labeled data for supervised learning, the generative adversarial net (GAN) framework~\cite{goodfellow2014generative} usually is adopted for handling this task~\cite{Perarnau2016,larsen2015autoencoding,zhou2017genegan, shen2017learning}.
%
%
%
However, visible artifacts and over-smoothing usually are inevitable in the transfer result for these methods.

\begin{figure*}[t]
\begin{center}
\includegraphics[width=0.9\linewidth]{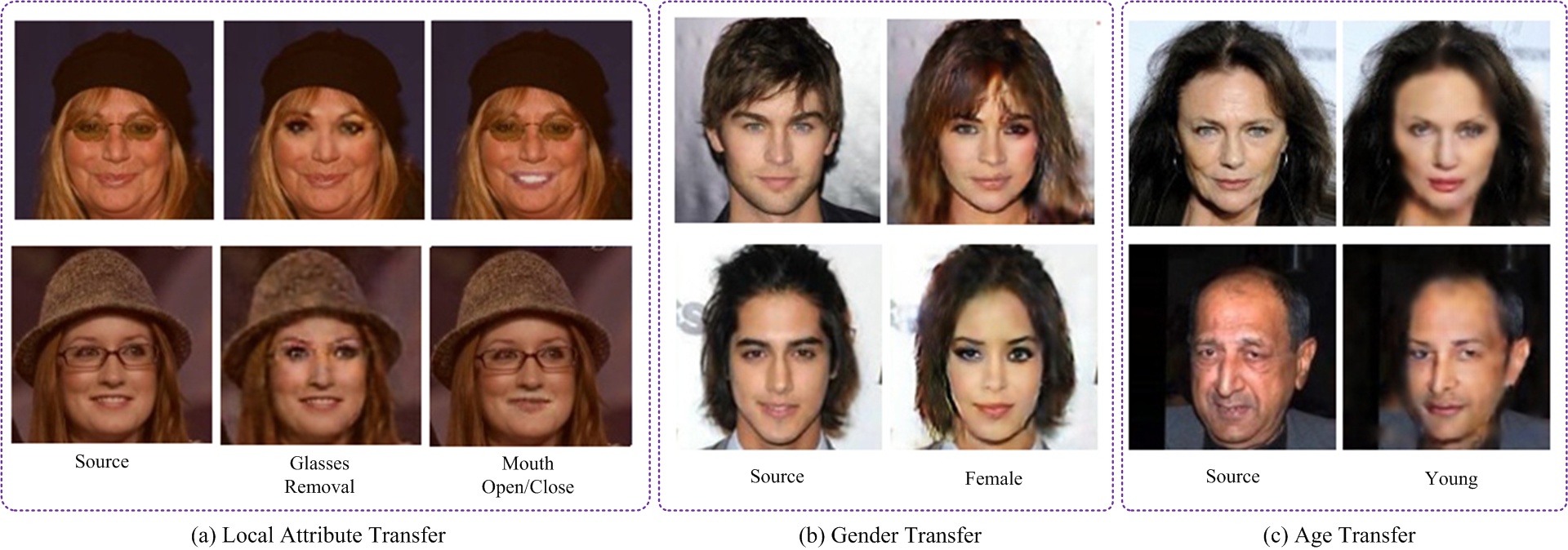}
\end{center}
   \caption{Illustration of the results by our DIAT on several facial attribute transfer tasks, including \emph{glasses removal}, \emph{mouth open/close}, \emph{gender transfer} and \emph{age transfer}. }
\label{outline}
\end{figure*}

In this paper, we present a novel Deep CNN model for Identity-Aware Transfer (DIAT) of facial attributes which can provide a unified solution to several facial animation and manipulation tasks, \eg, expression transfer, accessory removal, age progression, and gender transfer.
%
%
For each reference attribute label, we train a CNN model for the transfer of the input image to the desired attribute.
%
%
Note that the reference attribute may be only related to some parts of the image.
To avoid the the incorrect editing on attribute irrelevant region, our model consists of a mask network and an attribute transform network.
The attribute transform network is presented to edit the input image for generating the desired attribute.
While the mask network is adopted to estimate a mask of the attribute relevant region for guiding the combination of the input image and transformed image.
Then attribute transform network and mask network work collaboratively to generate final photo-realistic transfer result.

For attribute transfer, the ground truth transfer results generally are very difficult or even impossible to obtain. Therefore, we follow the GAN framework to train the model.
As for training data, we only consider the binary attribute labels presented in the large-scale CelebFaces Attributes (CelebA) dataset~\cite{liu2015faceattributes}.
To capture the convolutional feature distribution of each attribute, we construct an attribute guided set using all the images with the desired attribute in CelebA.
Then, the input set is defined as a set of input images without the reference attribute.
Due to the infeasibility of ground truth transfer results, two alternative losses, \ie, adversarial attribute loss and identity-aware perceptual loss, are incorporated for unsupervised training of our DIAT model.
Furthermore, two regularizers, \ie, perceptual regularization and attribute ratio regularization, are also introduced in the learning of DIAT.

In terms of attribute transfer, the generated image should have the desired attribute label.
%
%
Following the GAN framework, we define the adversarial attribute loss on the attribute discriminator to require the generated image to have the desired attribute.
As for identity-aware transfer, our DIAT requires that the generated image should keep the same or similar identity to the input image.
To this end, the identity-aware perceptual loss is introduced on the convolutional feature map of a CNN model to model the content similarity between the reference face and the generated face.
Instead of adopting any pre-trained CNNs, we suggest to define the perceptual loss on the attribute discriminator, which can be adaptively trained along with the learning procedure, and is named as adaptive perceptual loss.
Compared with conventional perceptual loss, ours is more effective in computation, tailored to our specific attribute transfer task, and can serve as a kind of hidden-layer supervision \cite{lee2015deeply} or regularization to ease the training of the DIAT model.
To further encourage the identity keeping property, we add an identity loss by minimizing the distance between feature representations of the generated face and the reference. Pre-trained VGG-Face is used to extract the identity related features for face verification. 
%

%
The model objective of DIAT also consider two regularizers, \ie, perceptual regularization and attribute ratio regularization.
To suppress the artifacts, we propose a denoising network to serve for a perceptual regularization on the generated image.
%
%
To guide the learning of mask network, an attribute ratio regularization is introduced to constrain the size of attribute relevant region.
%
Finally, our DIAT model can be learned from training data by incorporating adversarial attribute loss, adaptive perceptual loss with perceptual regularization and attribute ratio regularization.

Extensive experiments are conducted on CelebA and real images from the website \emph{iStock}\footnote{https://www.istockphoto.com/hk}.
As illustrated in Fig.~\ref{outline}, our DIAT performs favorably in attribute transfer with minor or no modification on the identity of the input faces.
Even for some identity-related attributes (\eg, gender), our DIAT can obtain visually impressive transfer result while retaining most identity-relevant features.
Computational efficiency is also a prominent merit of our method. In the testing stage, our DIAT can process more than one hundred of images within one second.
Furthermore, our model can be extended to face hallucination, and is effective in generating photo-realistic high resolution images.
A preliminary report of this work is given in 2016~\cite{li2016deep}. To sum up, our contribution is three-fold:
\begin{enumerate}
\item A novel DIAT model is developed for facial attribute transfer.
For better preserving of attribute irrelevant feature, our model comprises a mask network and an attribute transform network, which collaborate to generate the transfer result and can be jointly learned from training data.
%
\item Adversarial attribute loss, adaptive perceptual loss, identity loss, perceptual regularization, and attribute ratio regularization are incorporated for training our DIAT model.
The adversarial attribute loss is adopted to make the transfer result exhibit the desired attribute, and the adaptive perceptual loss is defined on the discriminator for identity-aware transfer while improving training efficiency.
Moreover, perceptual regularization and attribute ratio regularization are further introduced for suppressing the artifacts and constraining the mask network.
%
\item Experimental results validate the effectiveness and efficiency of our method for identity-aware attribute transfer.
Our DIAT can be used for the transfer of either local (\eg, mouth), global (\eg, age progression) or identity-related (\eg, gender) attributes, and can be extended to face hallucination.
\end{enumerate}

The remainder of the paper is organized as follows. Section~\ref{sec:related} gives a brief survey on relevant work. Section~\ref{sec:method} describes the model and learning of our DIAT method. Section~\ref{sec:exp} reports the experimental results on facial attribute transfer and face hallucination. Finally, Section~\ref{sec:conclusion} ends this work with several concluding remarks.

\section{Related work}
\label{sec:related}

Deep convolutional neural networks (CNNs) not only have achieved unprecedented success in versatile high level vision problems~\cite{krizhevsky2012imagenet, Parkhi15, girshick2014rich, VQA, zhang2017visual}, but also exhibited their remarkable power in understanding, generating, and recovering images~\cite{goodfellow2014generative,zhu2015learning, mahendran2015understanding,zhang2017beyond,jin2017inverse, du2017stacked}. In this section, we focus on the task of facial attribute transfer, and briefly survey the CNN models for image generation and face generation.

\subsection{CNN for image generation}

Generative image modeling is a critical issue for image generation and many low level vision problems. Conventional sparse~\cite{elad2006image}, low rank~\cite{gu2016wnnm}, FRAME~\cite{zhu1998frame} and non-local similarity~\cite{buades2005nlm,dabov2007BM3D} based models usually are limited in capturing highly complex and long-range dependence between pixels. For better image modeling, a number of CNN-based methods have been proposed, including convolutional auto-encoder~\cite{kulkarni2015deep}, PixelCNN and PixelRNN~\cite{van2016pixel}, and they have been applied to image completion and generation.

Several CNN architectures have been developed for image generation. Fully convolutional networks can be trained in the supervised learning manner to generate an image from an input image~\cite{dong2014learning,long2015fully}.
The generative CNN model~\cite{dosovitskiy2015learning} stacks four convolution layers upon five fully connected layers to generate images from object description.
Kulkarni \etal~suggest the Deep Convolution Inverse Graphics Network (DC-IGN), which follows the variational autoencoder architecture~\cite{kingma2013auto} to transform the input image into different pose and lighting condition.
However, both generative CNN~\cite{dosovitskiy2015learning} and DC-IGN~\cite{kingma2013auto} require many labeled images in training.

To visualize and understand CNN features, several methods have been proposed to reconstruct images by inverting deep representation~\cite{mahendran2015understanding} or maximizing class score~\cite{simonyan2013deep}.
Subsequently, Gatys \etal~\cite{gatys2015style} suggest to combine content and style losses defined on deep representation on the off-the-shelf CNNs for artistic style transfer.
To improve the efficiency, alternative approaches have been proposed by substituting the iterative optimization procedure with pre-trained feed-forward CNN~\cite{johnson2016perceptual, ulyanov2016texture}.
%
%
And perceptual loss has also been adopted for style transfer and other generation tasks~\cite{johnson2016perceptual}.
Motivated by these works, both identity-aware adaptive perceptual loss and perceptual regularization are exploited in our DIAT model to meet the requirement of facial attribute transfer.

Another representative approach is generative adversarial network (GAN), where a discriminator and a generator are alternatingly trained as an adversarial game~\cite{goodfellow2014generative}.
The generator aims to generate images to match the data distribution, while the discriminator attempts to distinguish between the generated images and the training data.
Laplacian Pyramid of GANs is further suggested to generate high quality image in a coarse-to-fine manner~\cite{denton2015deep}.
Radford \etal~\cite{radford2015unsupervised} extend GAN with the fully deep convolutional networks (\ie, DCGAN) for image generation.
To learn disentangled representations, information-theoretic extension of GAN is proposed by maximizing the mutual information between a subset of noise variables and the generated results~\cite{chen2016infogan}.
In~\cite{arjovsky2017wasserstein,gulrajani2017improved}, WGAN and WGAN-GP minimize the Wasserstein-1 distance between the generated distribution and the real distribution to improve the stability of learning generator.
%
%
In this work, we adopt the WGAN framework to learn our DIAT model, and further suggest adaptive perceptual loss for identity-aware transfer and perceptual regularization to suppress visual artifacts.
\begin{figure*}[t]
\begin{center}
   \includegraphics[width=1.0\linewidth]{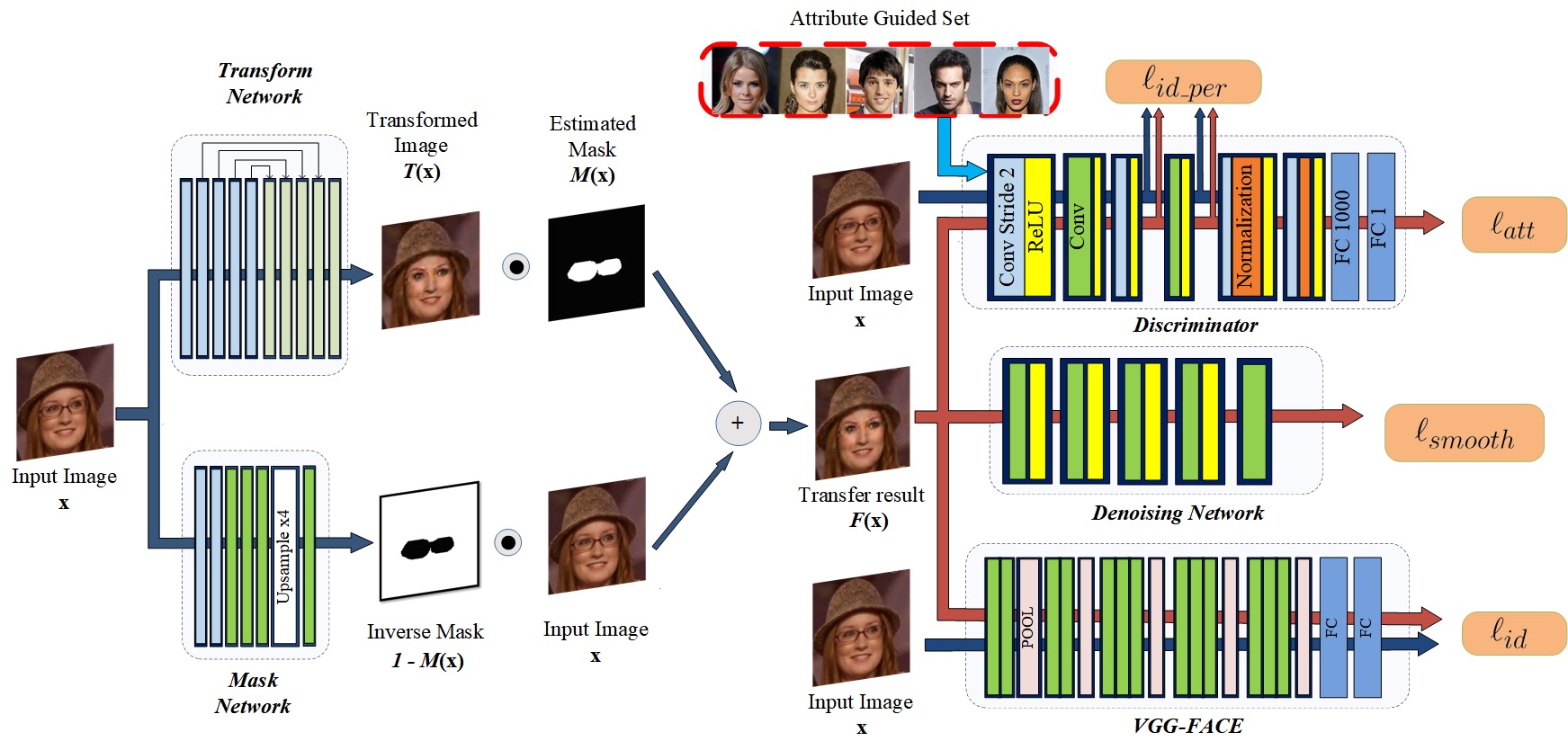}
\end{center}
   \caption{Schematic illustration of our DIAT model. Here we use \emph{glasses removal} as an example. The whole attribute transfer network $F(\mathbf{x})$ includes two sub-networks, \ie, a mask network to find the attribute relevant region $M(\mathbf{x})$ and an attribute transform network to produce the transformed image $T(\mathbf{x})$. Then $M(\mathbf{x})$ and $T(\mathbf{x})$ collaborate to generate the transfer result $F(\mathbf{x})=M(\mathbf{x}) \circ T(\mathbf{x})+(1-M(\mathbf{x})) \circ \mathbf{x}$. In order to learn $F(\mathbf{x})$ from training data, we incorporate adversarial attribute loss $\ell_{att}$, identity loss $\ell_{id}$, adaptive perceptual loss $\ell_{id\_per}$ with perceptual regularization $\ell_{smooth}$. Besides, an attribute ratio regularization is also adopted to constrain the estimated mask $M(\mathbf{x})$.}
\label{diat_procedure}
\end{figure*}

\subsection{CNN for face generation}

Facial attribute transfer has received considerable recent attention.
Larsen \etal~\cite{larsen2015autoencoding} present to combine variational autoencode with GAN (VAE/GAN) for image generation.
By modeling the attribute vector as the difference between the mean latent representations of the images with and without the reference attribute, VAE/GAN can provide a flexible solution to arbitrary facial attribute transfer, but is limited in transfer performance.
Li \etal~\cite{li2016convolutional} suggest an attribute driven and identity-preserving face generation model by solving an optimization problems with perceptual loss, which is computationally expensive and cannot obtain high quality results.
Perarnau \etal~\cite{Perarnau2016} adopt an encoder-decoder architecture, where attribute transfer can be conducted by editing the latent representation.
Shen \etal~\cite{shen2017learning} learn the residual image in the GAN framework, and adopt dual learning to learn two reverse attribute transfer models simultaneously.
Zhou \etal~\cite{zhou2017genegan} propose a model to learn object transfiguration from two sets of unpaired images that have the opposite attribute. 
However, most existing methods cannot achieve high quality transfer results, and visible artifacts and over-smoothing usually are inevitable.
In comparison, our DIAT model can achieve much better transfer results than the competing methods \cite{li2016convolutional, yeh2016semantic, larsen2015autoencoding, Perarnau2016}.

Besides, CNNs have also been developed for other face generation tasks.
For painting style transfer of head portrait, Selim \etal~\cite{selim2016painting} modify the perceptual loss to balance the contribution of the input photograph and the aligned exemplar painting.
Gucluturk \etal~train a feed-forward CNN with perceptual loss for sketch inversion.
Yeh \etal \cite{yeh2016semantic} apply DCGAN to semantic face inpainting in an optimization manner.



\section{Deep CNNs for Identity-aware Attribute Transfer}
\label{sec:method}

In this section, we present our DIAT model for identity-aware transfer of facial attribute.
As illustrated in Fig. \ref{diat_procedure}, our model involves a mask network and an attribute transform network which collaborate to produce the transfer result.
To train our model, we incorporate the adversarial attribute loss, adaptive perceptual loss, perceptual regularization and attribute ratio regularization.
%

%

\subsection{Network architecture}\label{s2_1}

Most facial attributes, \eg, expression and accessory, are local-based and only related to part of facial image.
Even for global attributes such as age and gender, some parts, \eg, the background, should also keep the same with the source image.
For the sake of preserving attribute irrelevant feature, it is natural to only perform attribute transfer in image region related to specific attribute.
However, it is not a trivial issue to find the attribute relevant region.
One possible solution is to manually specify the relevant region for each attribute given a new transfer task, but it undoubtedly restricts the universality and adaptivity of the solution.

In this work, we aim to provide a unified solution to attribute transfer, which indicates that we only require to prepare training data and retrain the model when a new transfer task comes.
To this end, our whole attribute transfer network is comprised of two sub-networks, \ie, mask network and attribute transform network.
Both mask network and attribute transform network take the source image $\mathbf{x}$ as input.
The mask network is utilized to predict a mask $M(\mathbf{x})$ to indicate the attribute relevant region, while the attribute transform network is used to produce the transformed image $T(\mathbf{x})$.
Given $M(\mathbf{x})$ and $T(\mathbf{x})$, the final transfer result can be obtained by,
\begin{equation}
\label{eqn:transfer}
F(\mathbf{x})=M(\mathbf{x}) \circ T(\mathbf{x})+(1-M(\mathbf{x})) \circ \mathbf{x},
\end{equation}
where $\circ$ denotes the element-wise product operator.
We also note that both attribute transform network and mask network can be learned from training data in an end-to-end manner.
In the following, we describe the architecture of attribute transform network and mask network, respectively.

\textbf{Attribute transform network.}
We adopt the Unet~\cite{ronneberger2015u} for attribute transform due to its good tradeoff between efficiency and reconstruction ability.
In general, the Unet architecture involves an encoder subnetwork and a decoder subnetwork, then skip connection and pooling operation are further introduced to exploit multi-scale information.
As for attribute transform, we design a 10-layer Unet, which includes 5 convolution layers for encoding and another 5 convolution layers for decoding.
In the encoder, we use convolution with stride 2 for downsampling.
In the decoder, a depth to width (DTOW) layer~\cite{shi2016real} is deployed for upsampling, and the element-wise summation operation is adopted to fuse the feature maps from the encoder and decoder subnetworks.
The detailed parameters of the attribute transform network are summarized in Table~\ref{table:transfer}.


\begin{table}[htb]
\caption{Architecture of the attribute transform network.}
\footnotesize
\begin{center}
\begin{tabular}{l|c}
\hline
Layer&  Activation size  \\
\hline
Input& $3\times128\times128$ \\
conv1, $4\times4\times64$, pad $1$, stride $2$ & $64\times64\times64$ \\
conv2, $4\times4\times128$, pad $1$, stride $2$ & $128\times32\times32$ \\
conv3, $4\times4\times256$, pad $1$, stride $2$ & $256\times16\times16$ \\
conv4, $4\times4\times512$, pad $1$, stride $2$ & $512\times8\times8$ \\
conv5, $4\times4\times512$, pad $1$, stride $2$ & $512\times4\times4$ \\
DTOW, stride 2 & $128\times8\times8$\\
conv6, $3\times3\times512$, pad $1$, stride $1$ & $512\times8\times8$ \\
Element-wise add, conv6 and conv4 & $512\times8\times8$\\
DTOW, stride 2 & $128\times16\times16$\\
conv7, $3\times3\times256$, pad $1$, stride $1$ & $256\times16\times16$ \\
Element-wise add, conv7 and conv3 & $256\times16\times16$\\
DTOW, stride 2 & $64\times32\times32$\\
conv8, $3\times3\times128$, pad $1$, stride $1$ & $128\times32\times32$ \\
Element-wise add, conv8 and conv2 & $128\times32\times32$\\
DTOW, stride 2 & $32\times64\times64$\\
conv9, $3\times3\times64$, pad $1$, stride $1$ & $64\times64\times64$ \\
Element-wise add, conv9 and conv1 & $64\times64\times64$\\
DTOW, stride 2 & $16\times128\times128$\\
conv10, $5\times5\times3$, pad $2$, stride $1$ &
$3\times128\times128$\\
\hline
\end{tabular}
\end{center}
\label{table:transfer}
\end{table}

\textbf{Mask network.}
As for mask network, we first adopt a 5-layer fully convolutional network to generate a $32 \times 32$ binary mask for indicating the attribute relevant region.
A batch normalization layer is added after each convolution layer.
Then, $4\times$ upsampling is deployed by simply replicating each element in the $32 \times 32$ binary mask $4\times 4$ times.
In order to make the generated image smooth, we further utilize a $5\times5$ Gaussian filter with the standard deviation of $1.6$ to produce the final mask $M(\mathbf{x})$.
To sum up, the details of the mask network are given in Table~\ref{table:mask}.



\begin{table}[htb]
\footnotesize
\caption{Architecture of the mask network.}
\begin{center}
\begin{tabular}{l|c}
\hline
Layer&  Activation size  \\
\hline
Input& $3\times128\times128$ \\
conv, $4\times4\times32$, pad $1$, stride $2$ & $32\times64\times64$ \\
conv, $4\times4\times64$, pad $1$, stride $2$ & $64\times32\times32$ \\
conv, $3\times3\times64$, pad $1$, stride $1$ & $64\times32\times32$ \\
conv, $3\times3\times64$, pad $1$, stride $1$ & $64\times32\times32$ \\
conv, $3\times3\times1$, pad $1$, stride $1$ & $1\times32\times32$ \\
binarization\\
4$\times$upsampling & $1\times128\times128$\\
conv, $5\times5\times1$, pad $2$, stride $1$ &
$1\times128\times128$\\
\hline
\end{tabular}
\end{center}
\label{table:mask}
\end{table}

In our mask network, ReLU is adopted for nonlinearity for the first 4 convolution layers.
As for the fifth convolution layer, we adopt the sigmoid nonlinearity, and the binarization operation is then used to obtain the binary mask,
\begin{equation}
\label{eqn:binarization}
B(e_{ijk}) = \begin{cases}
1, & \mbox{if } e_{ijk} > 0.5\\
0, & \mbox{if } e_{ijk} \leq 0.5
\end{cases}
\end{equation}
where $e_{ijk}$ denotes an element of the feature map.
However, the gradient of the binarizer $B(e_{ijk})$ is zero almost everywhere except that it is infinite when $e_{ijk} = 0.5$,
%
%
making any layer before the binarizer never be updated during training.
As a remedy, we follow the straight-through estimator on gradient~\cite{courbariaux2016binarized}, and introduce a piecewise linear proxy function $\tilde{B}(e_{ijk})$ to approximate $B(e_{ijk})$,
\begin{equation}
\label{eqn:approx_bin}
\tilde{B}(e_{ijk})=\begin{cases}
1, & \mbox{if } e_{ijk} > 1\\
e_{ijk}, & \mbox{if }  1\leq e_{ijk} \leq 0 \\
0, & \mbox{if }  e_{ijk} < 0
\end{cases}.
\end{equation}
During training, $B(e_{ijk})$ is still used in forward-propagation calculation, while $\tilde{B}(e_{ijk})$ is used in back-propagation, with its gradient computed by,
\begin{equation}
\label{eqn:grad_bin}
\tilde{B}^\prime(e_{ijk})=\begin{cases}
1, & \mbox{if } 1\leq e_{ijk}\leq 0 \\
0, & \mbox{otherwise}
\end{cases}.
\end{equation}

\subsection{Model objective}\label{s2_1_1}
By enforcing proper constraints on transfer result $F(\mathbf{x})$, both the attribute transform network and mask network can be learned from training data.
However, the ground truth of attribute transfer usually is unavailable and not unique.
For example, it is generally impossible to obtain the ground truth of gender transfer in reality.
Instead, the training data used in this work includes a guided set $\mathcal{X}_G$ of images with the desired reference attribute and a source set $\mathcal{X}_S$ of input images not with the reference attribute.
And we do not require the images from guided set to have the same identity with theose from source set.
%

For the sake of identity-aware attribute transfer, we define two alternative losses: 
(i) adversarial attribute loss to make the transfer result exhibit the desired attribute, and
(ii) adaptive perceptual loss and identity loss to make the generated image keep the same or similar identity to input image.
To suppress the visual artifacts of the transfer result, we further include (iii) a perceptual regularization defined on a denoising network.
Finally, (iv) an attribute ratio regularization is deployed on $\sum_{i,j,k} B(e_{ijk})$ to guide the learning of mask network.
In the following, we provide more details on these losses and regularization terms.


\textbf{Adversarial attribute loss.}\quad
%
%
Adversarial strategy is a common stratey that is widely used in security problems~\cite{garnaev2016security,zhang2016adversarial}. For computer vision, an adversarial learning framework called generative adversarial networks(GANs) also shows powerful ability on generating images that fulfil the distribution of a set of images without any ground truth targets. 
Considering the infeasibility of obtaining the ground truth transfer result, we define the adversarial attribute loss based on the guided set $\mathcal{X}_G$ and the source set $\mathcal{X}_S$.
%
%
%
%
If the guided set $\mathcal{X}_G$ is of large scale, it can provide a natural representation of the attribute distribution.
Therefore, the goal of adversarial attribute loss is to make that the distribution of generated images matches the real attribute distribution.
To this end, we adopt the generative adversarial network framework, where the generator is the attribute transfer network $F(\mathbf{x})$,
and the discriminator $D$ is used to define the adversarial attribute loss. 
The details of the discriminator are provided in Table~\ref{table:discriminator}, which contains 6 convolution layers followed by another two fully-connected layers.

\begin{table}[htb]
\footnotesize
\caption{Network architecture of the attribute discriminator.}
\begin{center}
\begin{tabular}{l|c}
\hline
Layer&  Activation size  \\
\hline
Input& $3\times128\times128$ \\
conv, $8\times8\times32$, pad $3$,  stride $2$ & $32\times64\times64$ \\
conv, $3\times3\times32$, pad $1$,  stride $1$ & $32\times64\times64$ \\
conv, $4\times4\times64$, pad $1$,  stride $2$ & $64\times32\times32$ \\
conv, $3\times3\times64$, pad $1$,  stride $1$ & $64\times32\times32$ \\
conv, $4\times4\times128$, pad $1$, stride $2$ & $128\times16\times16$ \\
conv, $4\times4\times128$, pad $1$, stride $2$ & $128\times8\times8$ \\
Fully connected layer with $1000$ hidden units& $1000$ \\
Fully connected layer with $1$ hidden units & $1$ \\
\hline
\end{tabular}
\end{center}
\label{table:discriminator}
\end{table}

Denote by $\mathbf{x}$ an input image from $\mathcal{X}_S$, and $\mathbf{a}$ an image from $\mathcal{X}_G$.
Let $p_{source}(\mathbf{x})$ be the distribution of the input images, $p_{att}(\mathbf{a})$ be the distribution of the images with the reference attribute.
The discriminator is defined as $D(\mathbf{a})$ to output the probability that the image $\mathbf{a}$ comes from the set $\mathcal{X}_G$.
To train the generator and the discriminator, we take use of the following adversarial attribute loss,
\begin{eqnarray}\label{gan}
  \min\limits_F\max\limits_D&&\mathbb{E}_{ \mathbf{a} \sim p_{att}(\mathbf{a})} \log D(\mathbf{a}) + \mathbb{E}_{\mathbf{x}\sim p_{source}(\mathbf{x})}\nonumber\\
  &&\log [1-D(F(\mathbf{x}))].
\end{eqnarray}
In order to improve the training stability, the improved Wasserstein GAN is adopted by defining the loss as,
\begin{eqnarray}\label{gan_wass}
  \min\limits_F\max\limits_D\mathbb{E}_{ \mathbf{a} \sim p_{att}(\mathbf{a})}[ D(\mathbf{a})]- \mathbb{E}_{\mathbf{x}\sim p_{source}(\mathbf{x})}
  [ D(F(\mathbf{x}))].
\end{eqnarray}
For simplicity, we respectively define the adversarial attribute losses for the generator and discriminator as follows,
\begin{eqnarray}\label{gan_generator}
  \min\limits_F  \ell_{att,F} = \{ - \mathbb{E}_{\mathbf{x}\sim p_{source}(\mathbf{x})}
  [ D(F(\mathbf{x}))]\},
\end{eqnarray}
\begin{eqnarray}\label{gan_disc}
  \min\limits_D  \ell_{att,D} \!=\!  \{\mathbb{E}_{\mathbf{x}\sim p_{source}(\mathbf{x})} [ D(F(\mathbf{x}))] - \mathbb{E}_{ \mathbf{a} \sim p_{att}(\mathbf{a})}[ D(\mathbf{a})] \}.
\end{eqnarray}

\textbf{Adaptive perceptual loss.}\quad  The adaptive perceptual loss is introduced to guarantee that the transfer result keeps the same or similar identity with the input image.
Due to identity is a high level semantic concept, it is not proper to define identity-aware loss by forcing two images to be exactly the same in pixel domain.
Instead, we define the squared-error loss on the feature representations of the discriminator, resulting in our adaptive perceptual loss.

Denote by $D$ the discriminator, and $D_l(\mathbf{x})$ the feature map of the $l$-th convolution layer.
$C_l$, $H_l$ and $W_l$ represent the channel number, height, and width of the feature map, respectively.
We then define the perceptual loss between $\mathbf{x}$ and $\hat{\mathbf{x}} = F(\mathbf{x})$ on the $l$-th convolution layer as,
\begin{equation}\label{eq10}
 \ell_{adaptive}^{D,l}(\hat{\mathbf{x}}, \mathbf{x}) = \frac{1}{2C_lH_lW_l}\|D_l(\hat{\mathbf{x}})-D_l(\mathbf{x})\|_F^2.
\end{equation}
And the identity-aware adaptive perceptual loss is further defined as,
\begin{equation}\label{eq11}
  \ell_{id\_{per}}(\mathbf{x}) = \sum_{l=3}^4 w_l\ell_{adaptive}^{D,l}(T(\mathbf{x}), \mathbf{x}).
\end{equation}

We note that the discriminator $D$ is learned from training data.
Thus, the network parameters of $\ell_{id\_per}(\mathbf{x})$ will be changed along with the updating of discriminator, and thus we name $\ell_{id\_per}(\mathbf{x})$ as adaptive perceptual loss.
In contrast, conventional perceptual loss~\cite{johnson2016perceptual} is defined on the off-the-shelf CNNs (\eg, VGG-Face~\cite{Parkhi15}).
Compared with conventional perceptual loss~\cite{johnson2016perceptual}, our adaptive perceptual loss generally is more effective in improving the training efficiency and attribute transfer performance:
\begin{enumerate}
\item The training efficiency of adaptive perceptual loss can be further explained from two aspects.
      (i) For conventional perceptual loss, the forward and backward calculations are required for both the off-the-shelf CNN and the discriminator during training.
      Due to that the adaptive perceptual loss is defined on the discriminator, it is sufficient to only conduct forward and backward calculation on the discriminator, making our DIAT more efficient in training.
      (ii) For conventional GAN, the generator usually is difficult to be trained.
      As for our DIAT, it can be trained by both the adversarial attribute loss and adaptive perceptual loss, greatly accelerating the training speed.
      Actually, the adaptive perceptual loss is defined on the third and fourth convolution layers of the discriminator, which can serve as some kind of hidden-layer supervision and benefit the convergence of network training \cite{lee2015deeply}.
\item For conventional perceptual loss, the off-the-shelf CNNs generally are pre-trained using other training data and are not tailored to attribute transfer.
      One plausible choice is the VGG-Face~\cite{Parkhi15}, which, however, is trained for face recognition and may not be suitable for identity-aware attribute transfer.
      In comparison, our adaptive perceptual loss is defined on the discriminator which is trained for modeling $p_{att}(\mathbf{a})$.
      Such loss can thus provide natural balance between identity similarity and attribute transfer and benefit transfer performance.
      For example, in terms of gender transfer, the introduction of adaptive perceptual loss will allow the adaptive adjustment on the length of hair.

\end{enumerate}

\textbf{Identity Loss.}\quad
The proposed adaptive perceptual loss does help keep the content similarity between the generated face and the reference. However, it cannot guarantee the identity by itself. To further enhance the identity keeping property, we add constrains on the feature representation extracted for face recognition~\cite{huo2018heterogeneous} or verification~\cite{zheng2018pairwise}. In face verification task, two faces are from the same person when the distance between two features are smaller than ceratain threshold. Here, we adopt VGG-Face and model the distance between the features of the generated face and the reference as the identity loss,

\begin{equation}\label{eq11_1}
  \ell_{id}(\mathbf{x}) = \|VGG(x)-VGG(T(\mathbf{x}))\|_2^2.
\end{equation}

\textbf{Perceptual regularization.}\quad
Despite the use of adversarial attribute loss and adaptive perceptual loss, visual artifacts are still inevitable in the transfer result.
Image regularization is thus required to encourage the spatial smoothness while preserving small scale details of the generated face $F(\mathbf{x})$.
One choice is the Total Variation (TV) regularizer which has been adopted in CNN feature visualization~\cite{mahendran2015understanding} and artistic style transfer~\cite{gatys2015style,johnson2016perceptual}.
However, the TV regularizer is limited in recovering small-scale texture details and suppressing complex artifacts.
Moreover, it is a generic model that does not consider the characteristics of facial images.
%
%

In this work, we take the facial characteristics into account and train a denoising network for perceptual regularization.
To train the denoising network, we generate the noisy image by adding Gaussian noise with the standard deviation of $15$ to the clean facial image from CelebA.
Inspired by residual learning~\cite{zhang2017beyond}, we train the denoising network through learning the residual between the noise image and the clean image.
Taking the noise image $\mathbf{y}$ as input, the denoising network utilizes a fully convolutional network of 6 layers to predict the residual $DN(\mathbf{y})$.
The denoising result can then be obtained by $\mathbf{y} + DN(\mathbf{y})$. The architecture of $DN(\mathbf{y})$ is listed in Table~\ref{table:denoising}.

\begin{table}[htb]
\footnotesize
\caption{Network architecture of the denoising network.}
\begin{center}
\begin{tabular}{l|c}
\hline
Layer&  Activation size  \\
\hline
Input& $3\times128\times128$ \\
conv, $3\times3\times64$, pad $1$,  stride $1$ & $64\times128\times128$ \\
conv, $3\times3\times64$, pad $1$,  stride $1$ & $64\times128\times128$ \\
conv, $3\times3\times64$, pad $1$,  stride $1$ & $64\times128\times128$ \\
conv, $3\times3\times64$, pad $1$,  stride $1$ & $64\times128\times128$ \\
conv, $3\times3\times64$, pad $1$,  stride $1$ & $64\times128\times128$ \\
conv, $3\times3\times3$, pad $1$,  stride $1$ & $3\times128\times128$ \\
\hline
\end{tabular}
\end{center}
\label{table:denoising}
\end{table}

Denote by $\mathcal{T} = \{(\mathbf{y}_i, \mathbf{x}_i)\}_{i=1}^{n}$ a training set, where $\mathbf{y}_i$ denotes the $i$-th noisy image and $\mathbf{x}_i$ the corresponding clean image.
The objective for learning $DN(\mathbf{y})$ is given as,
\begin{equation}
\min \|DN(\mathbf{y}) + \mathbf{y} - \mathbf{x}\|^2_F.
\end{equation}
Given the denoising network and the transfer result $F(\mathbf{x})$, we define the perceptual regularization as,
\begin{equation}
\label{eqn:regularization}
\ell_{smooth}(F(\mathbf{x})) = \max \{0, \|DN(F(\mathbf{x}))\|^2_F-t\},
\end{equation}
where $\|\cdot\|_F$ denotes the Frobenius norm.
Note that $DN(F(\mathbf{x}))$ predicts the residual between the latent clean image and $F(\mathbf{x})$. Minimizing $\|DN(F(\mathbf{x}))\|^2_F$ makes $F(\mathbf{x})$ be close to the clean image, and can be used to suppress the noise and artifacts in $F(\mathbf{x})$.
Furthermore, the threshold $t$ is introduced for better preserving of small scale details, and we empirically set $t$ be a value in the range of $[16, 30]$.
Note that the regularizer in Eqn. (\ref{eqn:regularization}) is defined on the denoising network $DN(F(\mathbf{x}))$, and thus is named as perceptual regularization.


%


\textbf{Attribute ratio regularization.}\quad
The size of attribute relevant region varies for different attributes.
For example, the region related to \emph{mouth open/close} mainly includes the mouth and should be small.
For \emph{glasses removal}, the attribute relevant region includes the two eyes and is relatively large.
As for \emph{gender transfer}, all the face region and the hair should be attribute relevant.
Therefore, we introduce an attribute ratio regularization term to constrain the size of attribute relevant region.
Specifically, such regularization is defined on the binary mask in Eqn. (\ref{eqn:binarization}).
Denote by $N$ the image size, and $p$ the expected ratio of the region for a specific attribute.
The attribute ratio regularization is then defined as,
\begin{equation}
\label{eqn:ratio_regularization}
\ell_{mask}(M(\mathbf{x})) = (\sum M(\mathbf{x})- pN)^2
\end{equation}
In our experiments, we set smaller $p$ value for local attribute and larger $p$ value for global attribute.

%

\textbf{Objective function.}\quad
We define the objective function for learning the transfer model $F$ and the discriminator $D$ by combining the adversarial attribute loss, adaptive perceptual loss, perceptual regularization, and attribute ratio regularization.
The transfer model $F(\mathbf{x}) = M(\mathbf{x}) \circ T(\mathbf{x})+(1-M(\mathbf{x})) \circ \mathbf{x}$ is learned by minimizing the following objective,
\begin{eqnarray}\label{eqn:objectiveF}
\min\limits_{M, T} \ell_{F}(\mathbf{x}) + \mu \ell_{smooth}(F(\mathbf{x})),
\end{eqnarray}
where $\ell_{F}(\mathbf{x}) = \ell_{att,F} + \lambda(\ell_{id}(\mathbf{x})+\ell_{id\_per}(\mathbf{x})) + \gamma \ell_{mask}(M(\mathbf{x}))$.
$\lambda$, $\gamma$, and $\mu$ are the tradeoff parameters for the adaptive perceptual loss, attribute ratio regularization, and perceptual regularization, respectively.
However, it is difficult to set the tradeoff parameter $\mu$.
Instead, we empirically find that the transfer model can be stably learned by alternatingly minimizing $\ell_{F}(\mathbf{x})$ and $\ell_{smooth}(F(\mathbf{x}))$ during training.
Finally, the discriminator $D$ is learned by minimizing the following objective,
\begin{eqnarray}\label{eq7}
\min\limits_{D} \!\!\!&&\!\!\! \ell_{att,D}.
\end{eqnarray}


\subsection{Learning algorithm}\label{sec:learning}

Generally, both the generator and the discriminator are difficult to converge in GAN.
Therefore, we adopt a two-stage strategy for learning the transfer model $F(\mathbf{x})$ and the discriminator $D$:
(i) we first combine the source set $\mathcal{X}_S$ and the guided set $\mathcal{X}_G$ to pre-train for initialization, and (ii) alternate between updating $F$ and $D$.
The procedure for training the transfer model $F(\mathbf{x})$ is summarized in Algorithm~\ref{alg1}.


\textbf{Initialization.}\quad
For the initialization of the $F$, we only consider the attribute transform network $T$, and leave the mask network $M$ be learned in the second stage.
Note that the transform network $T$ has the architecture of auto-encoder.
Thus, it can be pre-trained by minimizing the following reconstruction objective on $\mathcal{X}_S$ and $\mathcal{X}_G$,
\begin{equation}\label{eq8}
  \ell_{rec} = \sum_{\mathbf{x} \in \mathcal{X}_S \cup \mathcal{X}_G}||\mathbf{x} - T(\mathbf{x})||_F^2 .
\end{equation}

As for the initialization of the discriminator, we use the images in $\mathcal{X}_S$ as negative samples and the images in $\mathcal{X}_G$ as positive samples.
Then the discriminator can be pre-trained by minimizing the following objective,
\begin{equation}\label{eq9}
  \ell_{dis} = \sum_{\mathbf{x}_i \in \mathcal{X}_S \cup \mathcal{X}_G}||y_i-D(\mathbf{x}_i)||^2,
\end{equation}
where $y_i$ is $1$ for positive image $\mathbf{x}_i$ and $-1$ for negative image.
%
%
By this way, the initialization can provide a good start point and benefit the convergence and stability of DIAT training.

\textbf{Network training.}\quad
After the initialization of $T$ and $D$, network training is further performed by updating the whole $F$ (including both $T$ and $M$) and $D$ alternatingly.
Moreover, $F$ is updated by first $tstep$ iterations for minimizing $\ell_F$ and then $nstep$ iterations for minimizing $\ell_{smooth}$.
We apply the RMSProp solver~\cite{hinton2012rmsprop} to train the transfer network $F$ and the discriminator $D$ with a learning rate of $5\times 10^{-5}$.

\begin{algorithm}[!tbp]
\caption{Learning the attribute transfer network}\label{alg1}
\begin{algorithmic}[1]
\renewcommand{\algorithmicrequire}{\textbf{Input:}}
\renewcommand{\algorithmicensure}{\textbf{End}}
\REQUIRE Source set $\mathcal{X}_S$, guided set $\mathcal{X}_G$, $dstep=12$, $tstep=12$, $nstep=6$. mini-batch size is $50$.
\renewcommand{\algorithmicrequire}{\textbf{Output:}}
\renewcommand{\algorithmicensure}{\textbf{End}}
\REQUIRE The attribute transfer network $F$
\STATE
Pre-train the transform model $T$ by minimizing the objective in Eqn.~\eqref{eq8}.
\STATE
Pre-train the discriminator $D$ by minimizing the objective in Eqn.~\eqref{eq9}.
\WHILE {not converged}
\STATE
Select a mini-bath $\mathbf{X}$ from $\mathcal{X}_S$ to generate the transfer results, which is further combined with another mini-batch $\mathbf{A}$ from $\mathcal{X}_G$ to form the set $\mathbf{X}^\prime$ for training the discriminator. Here we set $|\mathcal{X}^\prime|=100$.
%
%
\FOR{$i=1\; to\; dstep$}
\STATE
Use the RMSProp solver to update the discriminator $D$ with Eqn.~\eqref{eq7} using the mini-batch $\mathbf{X}^\prime$.
\ENDFOR
\STATE
Clip the parameters of the discriminator. \\$D\leftarrow clip(D,-c,c)$
\FOR{$i=1\; to\; tstep$}
\STATE
Use the RMSProp solver to update $F$ by minimizing $\ell_F$ in Eqn.~\eqref{eqn:objectiveF}.
\ENDFOR

\FOR{$i=1\; to\; nstep$}
\STATE
Use the RMSProp solver to update $F$ by minimizing $\ell_{smooth}$ in Eqn.~\eqref{eqn:objectiveF}.
\ENDFOR
\ENDWHILE
\STATE
Return the transfer network $F$
\end{algorithmic}
\end{algorithm}

\begin{figure*}
\begin{center}
\includegraphics[width=0.9\linewidth]{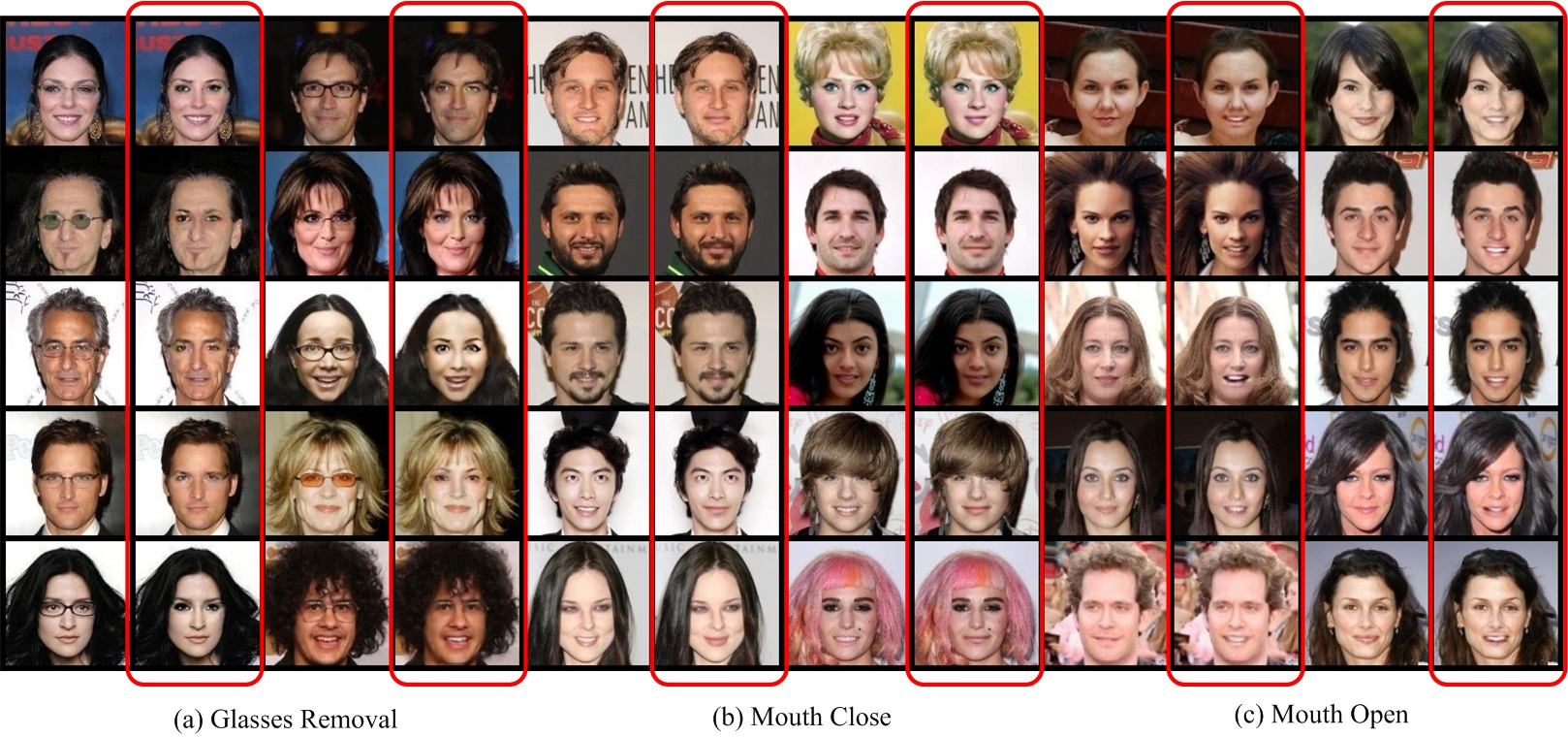}
\end{center}
   \caption{The results of local attribute transfer. For each task, the left and right columns are the input facial images and the transfer results, respectively.}
\label{local_results}
\end{figure*}

\subsection{Extension to face hallucination}\label{sec:facehallucinationmodel}
Besides facial attribute transfer, our DIAT can also be extended to other face editing tasks.
Here we use the $8\times$ face hallucination as an example.
Face hallucination is undoubtedly a global transfer task, and thus we remove the mask network $M$ as well as the attribute ratio regularization $\ell_{mask}$, making $F(\mathbf{x}) = T(\mathbf{x})$.
Moreover, the input image in face hallucination is of low resolution (LR) while the output image is of high resolution (HR).
To be consistent with attribute transfer, we super-resolve LR image to the size of HR image with the bicubic interpolator, which is taken as input to $F(\mathbf{x})$.

Furthermore, the ground truth HR images can be available to guide the network training for face hallucination.
Denote by $\mathbf{x}$ the super-resolved image by bicubic interpolator, and $\mathbf{y}$ the ground truth HR image.
Then, the pixel-wise reconstruction loss is defined as,
\begin{equation}
\label{eqn:rec_loss}
\ell_{rec}(\mathbf{x})= \|F(\mathbf{x})-\mathbf{y}\|_F^2.
\end{equation}
We further modify the definition of $\ell_F$ by removing the attribute ratio regularization and adding reconstruction loss,
\begin{equation}
\label{eqn:lossF_hall}
\ell_{F}(\mathbf{x}) = \ell_{att,F} + \lambda\ell_{id}(\mathbf{x}) + \beta \ell_{rec}(\mathbf{x}).
\end{equation}
where $\beta$ is the tradeoff parameter for pixel-wise reconstruction loss, and we set $\beta = 0.01$ in our experiment.
Given the training data, the models can then be learned by updating $F$ and $D$ alternatingly.

\section{Experimental Results}
\label{sec:exp}

%
In this section, we first describe the experimental settings, including the training and testing data, competing methods, model and learning parameters.
Experiments are then performed for local and global attribute transfer.
Quantitative metrics and the results on real images are also reported.
Moreover, we analyze the effect of adaptive perceptual loss and perceptual regularization.
Finally, the results are reported to further assess the performance of our DIAT for face hallucination.
The source code will be given after the publication of this work.

\begin{figure*}
\begin{center}
\includegraphics[width=0.9\linewidth]{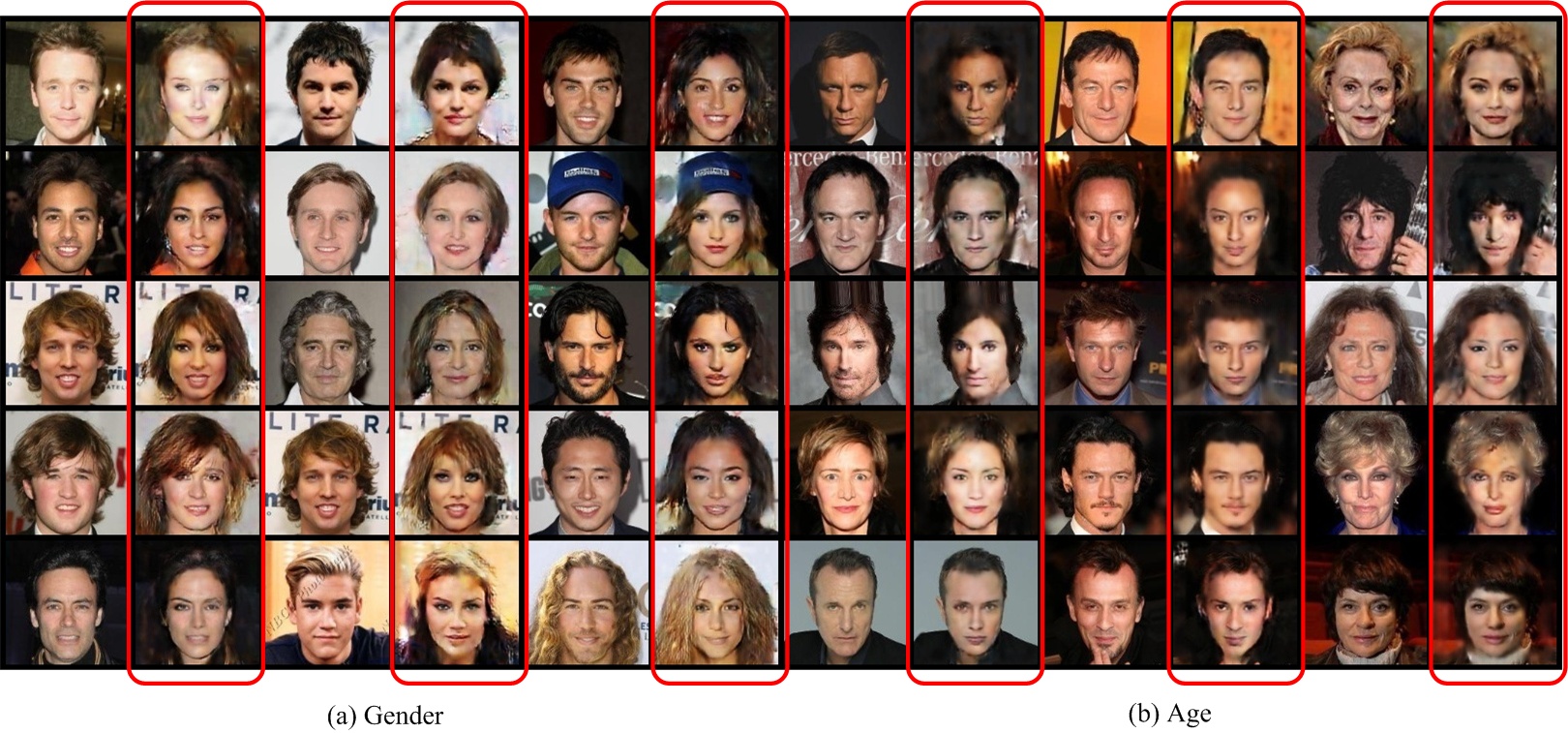}
\end{center}
   \caption{The results of global attribute transfer. For each task, the left and right columns are the input facial images and the transfer results, respectively.}
\label{global_results}
\end{figure*}

\subsection{Experimental settings}
Our DIAT models are trained using a subset of the aligned CelebA dataset~\cite{liu2015faceattributes} by removing the images with poor quality.
The size of the aligned images is $178\times218$. Due to the limitation of the GPU memory, we sample the central part of each image and resize it to $128\times128$.
%
%
For each attribute transfer task, we use all the images with the reference attribute from training set to form the guided set $\mathcal{X}_G$, and randomly select 10,000 training images not with the reference attribute as the source set $\mathcal{X}_S$.
After training, 2,000 images apart from the images for training are adopted to assess the attribute transfer performance.
And we also test the models on other real images from the website \emph{iStock}.

%
Only a few methods have been proposed for facial attribute transfer.
In our experiments, we compare our DIAT with the convolutional attribute-driven and identity-preserving model (CNIA)~\cite{li2016convolutional}, IcGAN~\cite{Perarnau2016} and VAE/GAN~\cite{larsen2015autoencoding} due to that their codes are available.
As for IcGAN and VAE/GAN, the original image size is not the same with our DIAT, so we resize the result to the same size with DIAT for comparison.
For the task of \emph{glasses removal}, we can first manually detect the region of glasses, and then use some face inpainting methods (\eg, semantic inpainting~\cite{yeh2016semantic}) to recover the missing pixels.
Thus we also compare our DIAT with semantic inpainting~\cite{yeh2016semantic} for \emph{glasses removal}.

All the experiments are conducted on a computer with the GTX TitanX GPU of 12GB memory.
We set the parameters $\lambda=1$ and $\gamma=0.01$ for DIAT.
For the threshold $t$ in the perceptual regularization $\ell_{smooth}$, we set it to be a value in the range of $[16, 30]$.
As for $p$ in the attribute ration regularization, we set it to be (i) $p=0.16$ for small local attributes (\eg, mouth), (ii) $p=0.32$ for large local attributes (\eg, eyes), and (iii) $p=0.62$ for global attributes (\eg, gender and age).
%

\subsection{Local attribute transfer}
\begin{figure*}
\begin{center}
\includegraphics[width=0.9\linewidth]{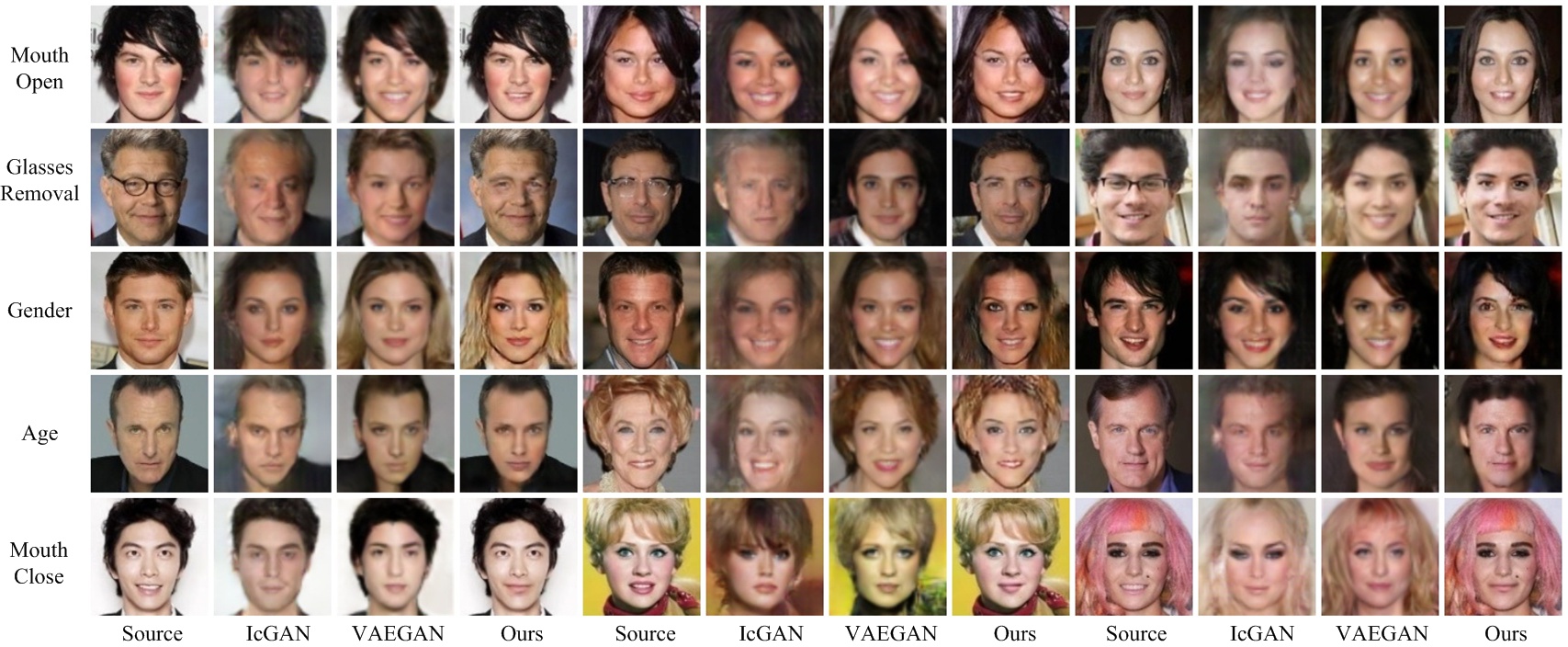}
\end{center}
   \caption{Comparison of transfer results by our DIAT, IcGAN~\cite{Perarnau2016} and VAE/GAN~\cite{larsen2015autoencoding}.}
\label{cmp_v2}
\end{figure*}

We assess the local attribute transfer models on three tasks, \ie, \emph{mouth open}, \emph{mouth close}, and \emph{eyeglasses removal}.
Fig.~\ref{local_results} illustrates the transfer results by our DIAT.
It can be seen that our DIAT performs favorably for transferring the input images to the desired attribute with satisfying visual quality.
Benefited from the mask network, the results by DIAT can preserve more identity-aware and attribute irrelevant details.
Moreover, when the training data are sufficient, it is feasible to separately train two DIAT models for reverse tasks, \eg, one for \emph{mouth open} and another for \emph{mouth close}.

\begin{figure*}
\begin{center}
\includegraphics[width=0.9\linewidth]{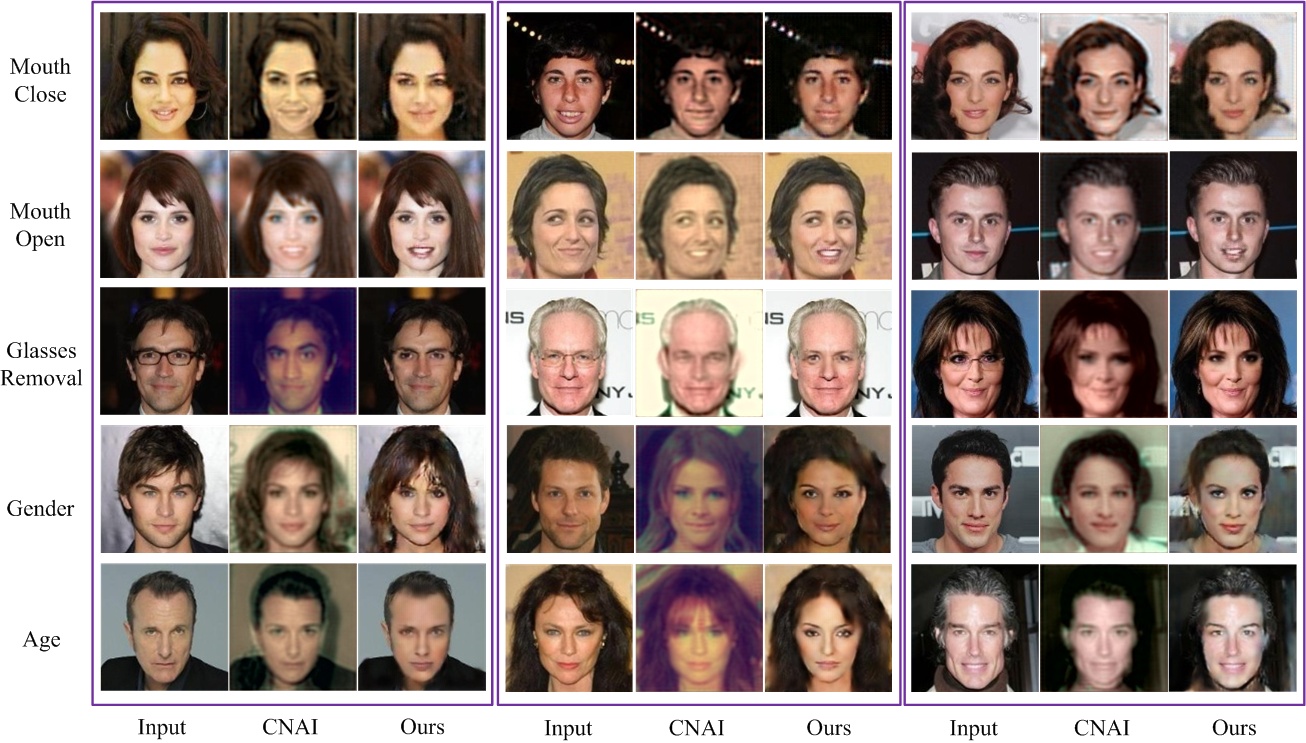}
\end{center}
   \caption{Comparison of transfer results by our DIAT and CNIA~\cite{li2016convolutional}.}
\label{cmp}
\end{figure*}

\begin{figure}[t]
\begin{center}
   \includegraphics[width=1.0\linewidth]{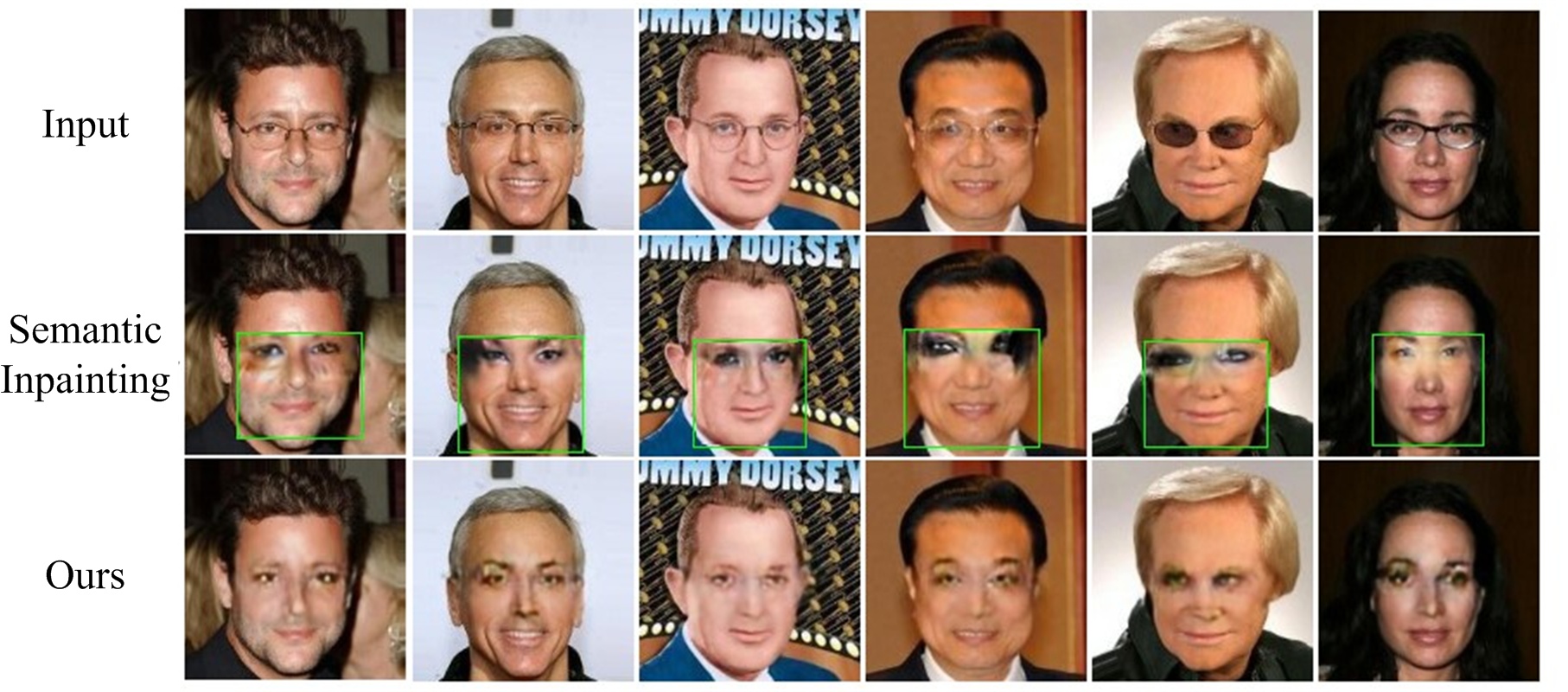}
\end{center}
   \caption{Comparison of the results by semantic inpainting~\cite{yeh2016semantic} for \emph{glasses removal}. The green rectangle shows the region of the input subimage for semantic inpainting}
\label{inpainting}
\end{figure}

We further compare our DIAT with three competing methods, \ie, CNIA~\cite{li2016convolutional}, IcGAN~\cite{Perarnau2016} and VAE/GAN~\cite{larsen2015autoencoding}.
As shown in Fig.~\ref{cmp}, the results by our DIAT are visually more pleasing than those by CNIA~\cite{li2016convolutional} for all the three local attribute transfer tasks.
In terms of run time, CNIA takes about $30$ seconds ($s$) to deal with an image, while our DIAT only needs $0.0045$ $s$.
Fig.~\ref{cmp_v2} further provides the results by IcGAN, VAE/GAN, and our DIAT.
In comparison with the competing methods, our DIAT can well address the attribute transfer tasks while recovering more visual details in both attribute relevant and attribute irrelevant regions.
Finally, for \emph{glasses removal}, we compare our DIAT with semantic inpainting~\cite{yeh2016semantic}, and the results in Fig.~\ref{inpainting} clearly demonstrate the superiority of DIAT.


\subsection{Global attribute transfer}

We consider two global attribute transfer tasks, \ie, \emph{gender transfer} and \emph{age transfer}.
For \emph{gender transfer}, we only evaluate the model for \emph{male-to-female}.
For \emph{age transfer}, we only test the model for \emph{older-to-younger}.
Fig.~\ref{global_results} shows the transfer results, and our DIAT is also effective for global attribute transfer.
Even \emph{gender transfer} certainly causes the change of the identity, as shown in Fig.~\ref{global_results}(a), our DIAT can still retain most identity-aware features, making the transfer result similar to the input image in appearance.
Figs.~\ref{cmp} and~\ref{cmp_v2} show the results by our DIAT, CNAI~\cite{li2016convolutional}, IcGAN and VAE/GAN.
Compared with the competing methods, the results by our DIAT well exhibit the desired attribute, and are of high visual quality with photo-realistic details.
Finally, we also note that for \emph{gender transfer} our DIAT is able of adjusting the hair length due to the introduction of adaptive perceptual loss.

%


\begin{figure*}
\begin{center}
\includegraphics[width=0.9\linewidth]{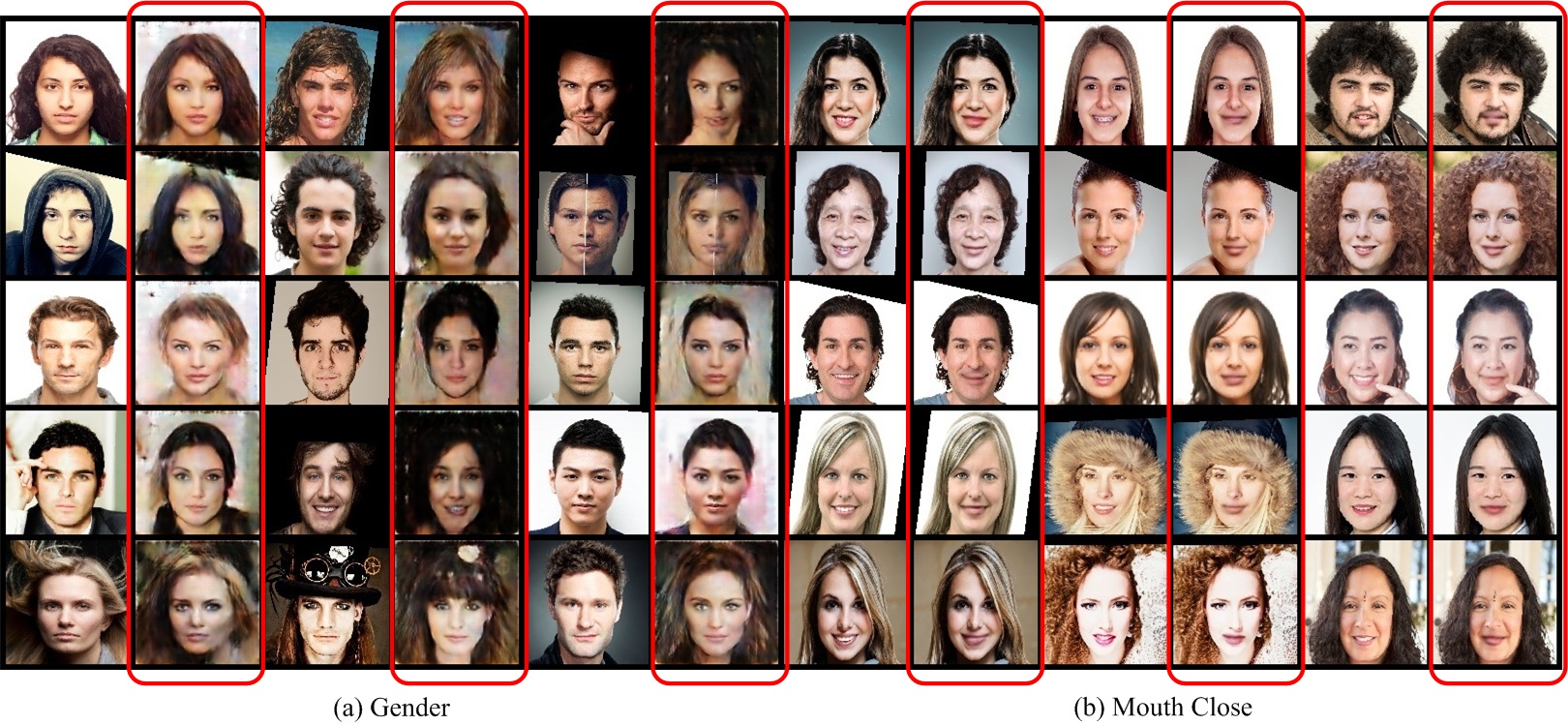}
\end{center}
   \caption{Local attribute transfer (\emph{mouth open}) and global attribute transfer (\emph{gender transfer}) on images from the website \emph{iStock}. For each task, the left and right columns are the input facial images and the transfer results, respectively.}
\label{other}
\end{figure*}

\subsection{Quantitative evaluation}
Given each attribute transfer task, we randomly select $2,000$ images without the reference attribute from the testing partition of CelebA to form our testing set.
Then, three groups of experiments are conducted to evaluate the transfer performance quantitatively:
\begin{itemize}
  \item \textbf{Attribute classification}. For attribute transfer, it is natural to require the transfer result to exhibit the desired attribute.
         Thus, we first train a CNN-based attribute classifier (including two convolution layers, three residual blocks and two fully-connected layers) using the training set of CelebA.
         Given an attribute transfer task, we test the classification accuracy of the desired attribute for the transfer results of 2,000 testing images.
         Tabel~\ref{table:classification} lists the classification accuracy for five attribute transfer tasks, \ie, \emph{mouth open}, \emph{mouth close}, \emph{glasses removal}, \emph{gender transfer}, and \emph{age transfer}.
         It can be observed that our DIAT achieves satisfying accuracy (\ie, $> 0.70$) for all the tasks, indicating that the results by our DIAT generally are with the desired attribute.
  \item \textbf{Identity verification}. As for local attribute transfer, we also require the transfer result to preserve the identity of input image.
         Here we use the open source face recognition platform \emph{Openface}\footnote{https://github.com/cmusatyalab/openface} for matching the input image with the transfer result.
         By setting the threshold be 0.99, Table~\ref{table:identification} lists the identity verification accuracy for \emph{mouth open}, \emph{mouth close}, and \emph{glasses removal}.
         The results demonstrate that our DIAT can well preserve the identity-aware feature for local attribute transfer.
  \item \textbf{Image quality}.
         Image quality is another crucial metric to assess attribute transfer.
         However, the ground truth of transfer result is unavailable, making it difficult to perform quantitative evaluation.
         Here we use a pair of reverse attribute transfer tasks (\ie, \emph{mouth close} and \emph{mouth open}) as an example, and adopt an indirect scheme to compute average PSNR on the 2000 testing images.
         Specifically, we first perform \emph{mouth open} to the images with \emph{mouth close}, and then perform reverse \emph{mouth close} to the transfer results.
         Finally, the input images are taken as ground truth and the images after two steps of transfer can be viewed as the generated images.
         By this way, we obtain the average PSNR of 33.27dB, indicating the effectiveness of our DIAT for local attribute transfer.
\end{itemize}

\begin{table}[htb]
\scriptsize
\caption{Attribute classification accuracy for transfer results.}
\begin{center}
\begin{tabular}{l|c|c|c|c}
\hline
mouth open & mouth close & glasses removal & gender & age  \\
\hline
0.821 & 0.806 & 0.763 & 0.684 & 0.702 \\
\hline
\end{tabular}
\end{center}
\label{table:classification}
\end{table}

\begin{table}[htb]
\scriptsize
\caption{Face verification accuracy for the transfer results.}
\begin{center}
\begin{tabular}{l|c|c}
\hline
mouth open & mouth close & glasses removal \\
\hline
0.912 & 0.903 & 0.872  \\
\hline
\end{tabular}
\end{center}
\label{table:identification}
\end{table}

\subsection{Results on other real facial images}
To assess the generalization ability, we use the DIAT models learned on CelebA to other real facial images from the website \emph{iStock}.
Each test image is first aligned with the 5 facial landmarks, and then input to the DIAT models.
Taking \emph{mouth open} and \emph{gender transfer} as examples, Fig.~\ref{other} gives the transfer results on 15 images for each task, clearly demonstrating the generalization ability of our models to other real facial images.

\begin{figure*}
\begin{center}
\includegraphics[width=0.9\linewidth]{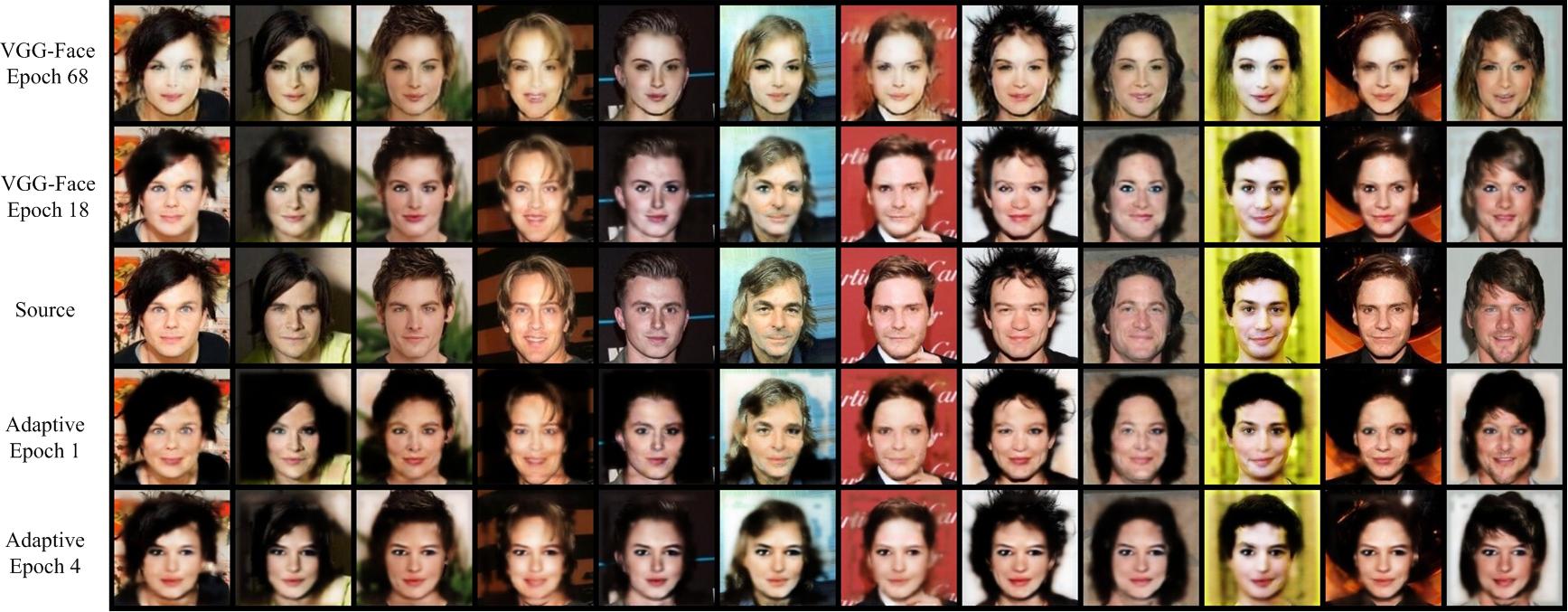}
\end{center}
   \caption{Comparison between the adaptive perceptual loss and the VGG-Face based perceptual loss.}
\label{perc}
\end{figure*}

\begin{figure}
\begin{center}
\includegraphics[width=1.0\linewidth]{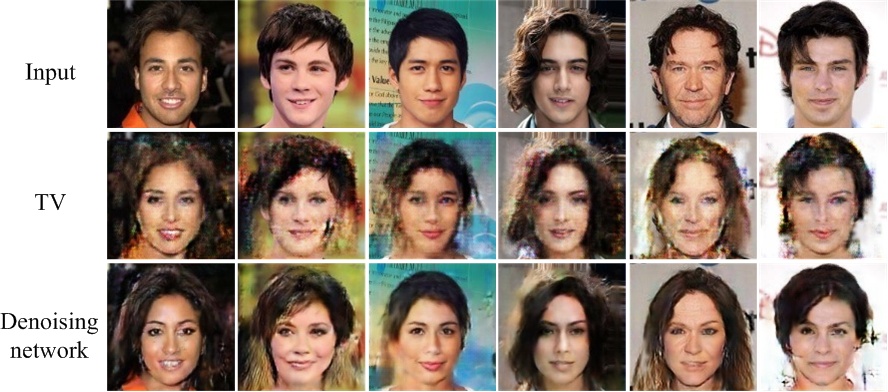}
\end{center}
   \caption{Comparison between the perceptual regularization and the TV regulization.}
\label{denoising}
\end{figure}

\subsection{Evaluation on adaptive perceptual loss and perceptual regularization}

We also implement a variant of DIAT (\ie, DIAT-1) by replacing the adaptive perceptual loss with the conventional perceptual loss defined on VGG-Face~\cite{Parkhi15}.
Taking gender transfer as an example, Fig.~\ref{perc} compares DIAT with DIAT-1.
It can be observed that DIAT converges very fast and can generate satisfying results after 4 epochs of training.
In comparison, DIAT-1 requires much more epochs in training, and the gender just begins to be modified after 18 epochs.
Moreover, the adoption of adaptive perceptual loss also benefits the transfer performance, and adaptive adjustment on the hair length can be observed on the transfer results by DIAT.
Furthermore, Fig.~\ref{denoising} shows the transfer results by DIAT with the perceptual regularization and the TV regularization.
It can be clearly seen that the perceptual regularization is more effective on suppressing noise and artifacts while preserving sharp edges and fine details.

\begin{table}[htb]
\footnotesize
\caption{Comparison of PSNR (in dB) for face hallucination.}
\begin{center}
\begin{tabular}{l|c|c|c}
\hline
 & Bicubic & Unet\cite{ronneberger2015unet} & DIAT  \\
\hline
PSNR& 29.68& 30.12 & 28.85  \\
\hline
SSIM& 0.606& 0.672 & 0.643  \\
\hline
\end{tabular}
\end{center}
\label{table:psnr}
\end{table}

\subsection{Results of the learnt mask}
\begin{figure}
\begin{center}
\includegraphics[width=0.9\linewidth]{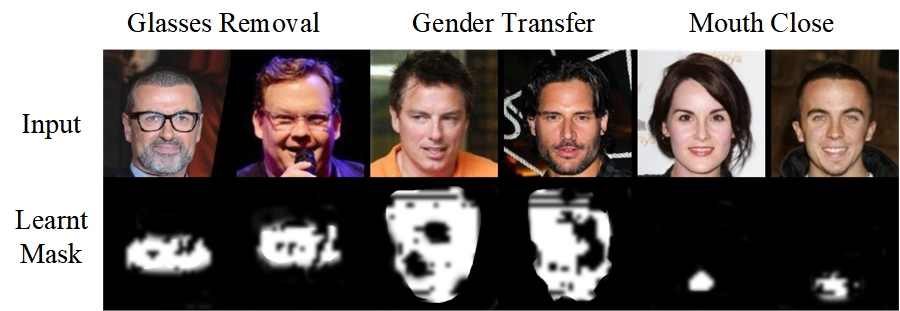}
\end{center}
   \caption{Results of the learnt mask of different transformation tasks.}
\label{mask}
\end{figure}

Fig.~\ref{mask} gives the masks generated by the mask network for different task. For the local attribute transformation tasks such as \textit{glasses removal} and \textit{closing mouth}, the generated masks accurately cover the local facial part which is related to the attribute. For global transformation like gender transformation, the mask covers most of the face and keep the background out.

\subsection{Experiments on face hallucination}

\begin{figure*}
\begin{center}
\includegraphics[width=0.9\linewidth]{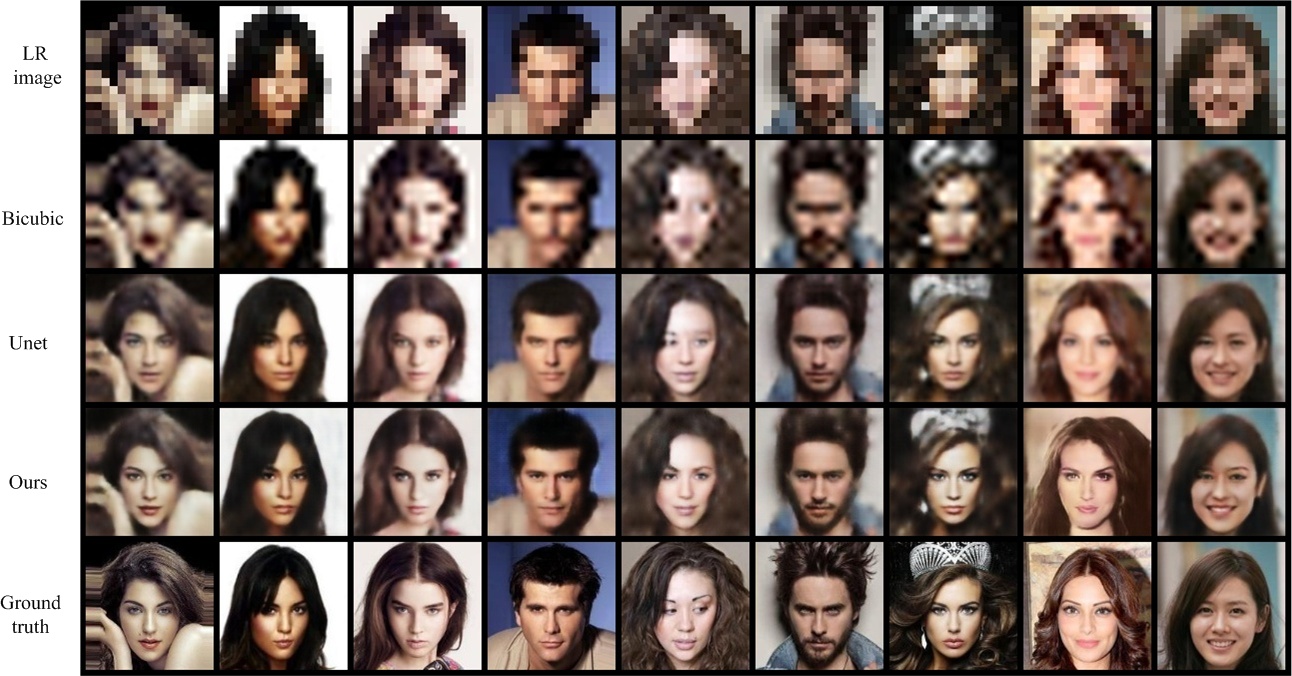}
\end{center}
   \caption{Results of face hallucination by different methods.}
\label{super}
\end{figure*}

Finally, we evaluate the performance of DIAT for $8\times$ face hallucination.
Table~\ref{table:psnr} lists the average PSNR and SSIM~\cite{wang2003ssim} values on the 2,000 testing images by DIAT, bicubic interpolator, and Unet, while Fig.~\ref{super} shows the super-resolved images.
Even our DIAT achieves lower PSNR/SSIM than the baseline Unet, it is much better in terms of visual quality, and can generate hallucinated image with rich textures and sharp edges.

%
%
%
%
%
%
%
%

\section{Conclusion}
~\label{sec:conclusion}

A deep identity-aware transfer (\ie, DIAT) model is presented for facial attribute transfer.
Considering that some attributes may be only related with parts of facial image, the whole transfer model consists of two subnetworks, \ie, mask network and attribute transform network, which work collaboratively to produce the transfer result.
In order to train the model, we further incorporate adversarial attribute loss, adaptive perceptual loss with perceptual regularization and attribute ratio regularization.
Experiments show that our model can obtain satisfying results for both local and global attribute transfer.
Even for some identity-related attributes (e.g., gender transfer), our DIAT can obtain visually impressive results with minor modification on identity-related features.
Our DIAT can also be extended to face hallucination and performs favorably in recovering facial details.
%
%
In future work, we will further improve the visual quality and diversity of the transfer results, and extend our model to arbitrary attribute transfer.


%

%
%
%
%
%

\ifCLASSOPTIONcaptionsoff
  \newpage
\fi



%

\bibliographystyle{IEEEtran}
\bibliography{IEEEabrv,database}

%








\end{document}